\documentclass{article}

\usepackage[preprint]{neurips_2021}

\usepackage[utf8]{inputenc} 
\usepackage[T1]{fontenc}    
\usepackage{hyperref}       
\usepackage{url}            
\usepackage{booktabs}       
\usepackage{amsfonts}       
\usepackage{nicefrac}       
\usepackage{microtype}      
\usepackage{xcolor}         

\usepackage{subfigure}

\usepackage{wrapfig}

\usepackage{cleveref}
\Crefformat{figure}{#2Fig.~#1#3}
\Crefmultiformat{figure}{Figs.~#2#1#3}{ and~#2#1#3}{, #2#1#3}{ and~#2#1#3}
\Crefformat{table}{#2Tab.~#1#3}
\Crefmultiformat{table}{Tabs.~#2#1#3}{ and~#2#1#3}{, #2#1#3}{ and~#2#1#3}

\usepackage{xspace}
\usepackage{graphicx}
\graphicspath{{figs/}}
\usepackage{caption}

\usepackage[frozencache=true,cachedir=.,newfloat]{minted}
\newcommand{\inputpython}[1]{
\inputminted[
    linenos,
    frame=lines,
    numbersep=-6pt,
    obeytabs=true
]{python}{#1}}
\newcommand{\inlinepython}[1]{\mintinline{python}{#1}}

\newcommand{\tyxe}{TyXe\xspace}
\newcommand{\pytorch}{Pytorch\xspace}
\newcommand{\pyro}{Pyro\xspace}

\title{\tyxe: Pyro-based Bayesian neural nets for \pytorch}

\author{%
  Facebook
}

\author{%
  Hippolyt Ritter\thanks{Work started at Uber AI Labs. HR is now at Facebook.}\\
  University College London\\
  \texttt{j.ritter@cs.ucl.ac.uk}\\
  \And
  Theofanis Karaletsos$^*$ \\
  Facebook \\
  \texttt{theokara@fb.com}
}

\begin{document}

\maketitle


\begin{abstract}
We introduce \tyxe, a Bayesian neural network library built on top of \pytorch and \pyro.
Our leading design principle is to cleanly separate architecture, prior, inference and likelihood specification, allowing for a flexible workflow where users can quickly iterate over combinations of these components.
In contrast to existing packages \tyxe does not implement any layer classes, and instead relies on architectures defined in generic \pytorch code.
\tyxe then provides modular choices for canonical priors, variational guides, inference techniques, and layer selections for a Bayesian treatment of the specified architecture.
Sampling tricks for variance reduction, such as local reparameterization or flipout, are implemented as effect handlers, which can be applied independently of other specifications.
We showcase the ease of use of \tyxe to explore Bayesian versions of popular models from various libraries: toy regression with a pure \pytorch neural network; large-scale image classification with torchvision ResNets; graph neural networks based on DGL; and Neural Radiance Fields built on top of \pytorch{3D}.
Finally, we provide convenient abstractions for variational continual learning.
In all cases the change from a deterministic to a Bayesian neural network comes with minimal modifications to existing code, offering a broad range of researchers and practitioners alike practical access to uncertainty estimation techniques.
The library is available at \url{https://github.com/TyXe-BDL/TyXe}.
\end{abstract}

\begin{listing*}[t]
\centering
\inputpython{snippets/bnn.py}
\vspace{-0.75em}
\caption{Bayesian nonlinear regression setup code example in 5 lines. Line $1$ is a standard \pytorch neural network definition, line $2$ is the likelihood of the data, corresponding to a data loss object. Line $3$ sets the prior and line $4$ constructs the approximate posterior distribution on the weights. Line $5$ finally brings all components together to set up the BNN. For MCMC, the \inlinepython{guide_factory} would be \inlinepython{HMC} or \inlinepython{NUTS} from \inlinepython{pyro.infer.mcmc} and the \inlinepython{BNN} a \inlinepython{tyxe.MCMC_BNN}.}
\label{lst:bnn}
\vspace{-1.25em}
\end{listing*}

\section{Introduction}
The surge of interest in deep learning has been fuelled by the availability of agile software packages that enable researchers and practitioners alike to quickly experiment with different architectures for their problem setting \citep{paszke2019pytorch, abadi2016tensorflow} by providing modular abstractions for automatic differentiation and gradient-based learning.
While there has been similarly growing interest in uncertainty estimation for deep neural networks, in particular following the Bayesian paradigm \citep{mackay1992practical,neal2012bayesian}, a comparable toolbox of software packages has mostly been missing.

A major barrier of entry for the use of Bayesian Neural Networks (BNNs) is the large overhead in required code and additional mathematical abstractions compared to stochastic maximum likelihood estimation as commonly performed in deep learning.
Moreover, BNNs typically have intractable posteriors, necessitating the use of various approximations when performing inference, which depending on the problem may perform better or worse and frequently require complex bespoke implementations.
This oftentimes leads to the development of inflexible small libraries or repetitive code creation that can lack essential ``tricks of the trade'' for performant BNNs, such as appropriate initialization schemes, gradient variance reduction \citep{kingma2015variational,tran2018simple}, or may only provide limited inference strategies to compare outcomes.
Even though various general purpose probabilistic programming packages have been built on top of those deep learning libraries (\pyro~\citep{bingham2019pyro} for \pytorch, Edward2~\citep{tran2018simple} for Tensorflow), software linking those to BNNs has only been released recently \citep{tran2019bayesian} and provides substitutes for Keras' layers~\citep{chollet2015keras} to construct BNNs from scratch.

In this work we describe \tyxe (Greek: chance), a package linking the expressive computational capabilities of \pytorch with the flexible model and inference design of \pyro \citep{bingham2019pyro} in service of providing a simple, agile, and useful abstraction for BNNs targeted at \pytorch practitioners.
Specifically, we highlight the following contributions we make through \tyxe:
\begin{itemize}
    \item We provide an intuitive, object-oriented interface that abstracts away \pyro to facilitate turning \pytorch-based neural networks into BNNs with minimal changes to existing code.
    \item Crucially, our design deviates from prior approaches, e.g. \citep{tran2019bayesian}, to avoid bespoke layer implementations, making \tyxe applicable to arbitrary \pytorch architectures.
    \item We make essential techniques for well-performing BNNs that are missing from \pyro, such as local reparameterization, available as flexible program transformations.
    \item \tyxe is compatible with architectures from libraries both native and non-native to the \pytorch ecosystem, such as torchvision ResNets and DGL graph neural networks.
    \item Leveraging \tyxe, we show that a Bayesian treatment of \pytorch{3d}-based Neural Radiance Fields improves their out-of-distribution robustness at a minimal coding overhead.
    \item Our modular design handily supports variational continual learning through updating the prior to the posterior. Such abstractions are also currently not available in \pyro.
\end{itemize}

In the following we give an overview of our library, with an initial focus on the API design followed by a range of research settings where \tyxe greatly simplifies `Bayesianizing' an existing workflow.
We provide experimental details in \Cref{app:experiments}, specifically discuss advancements upon \pyro in \Cref{app:comparison} and provide an in-depth overview of the codebase in \Cref{app:code}.



%

\section{\tyxe by example: non-linear regression}

The core components that users interact with in \tyxe are our BNN classes.
These wrap deterministic \pytorch \inlinepython{nn.Module} neural networks.
We then leverage \pyro to formulate a probabilistic model over the neural network parameters, in which we perform approximate inference.
There are two primary BNN classes with identical interfaces: \inlinepython{tyxe.VariationalBNN} and \inlinepython{tyxe.MCMC_BNN}.
Both offer a unified workflow of constructing a BNN, fitting it to data and then making predictions.
A more low-level class, \inlinepython{tyxe.PytorchBNN}, which can act as a drop-in BNN replacement for a \inlinepython{nn.Module} but lacks some of the high-level functionality of the other two classes, will be introduced in \Cref{sec:nerf}.
We stress that the former two classes only require using a \pyro optimizer in place of a \pytorch one, while the latter hides \pyro entirely, making its functionality accessible to \pytorch users without prior experience of using \pyro.

In this section we provide more details on each of the modelling steps along the example of a synthetic one-dimensional non-linear regression dataset.
We use the setup from \citep{foong2019between} with two clusters of inputs $x_1 \sim \mathcal{U}[-1, -0.7]$, $x_2 \sim \mathcal{U}[0.5, 1]$ and $y \sim \mathcal{N}(\cos(4x + 0.8), 0.1^2)$.

\subsection{Defining a BNN}

\begin{figure*}[t]
    \centering
    \subfigure[Local reparameterization]{\label{subfig:lr}
        \includegraphics[width=0.3\linewidth]{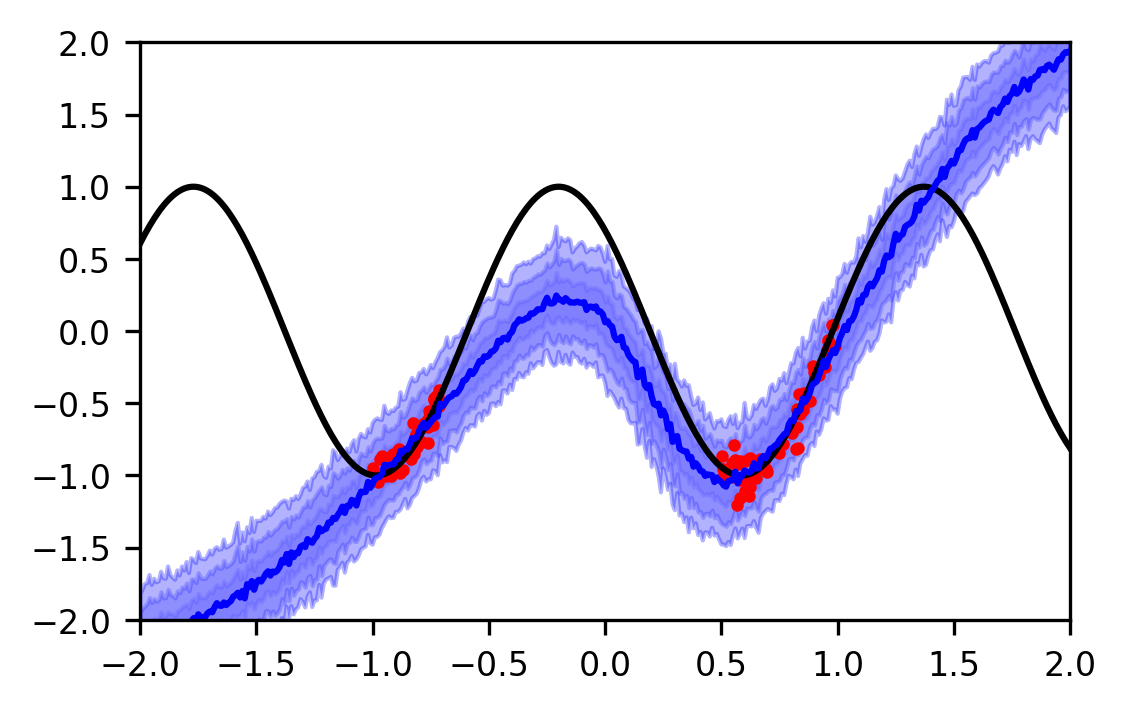}
    }%
    \hfill
    \subfigure[Shared weight samples]{\label{subfig:nolr}
        \includegraphics[width=0.3\linewidth]{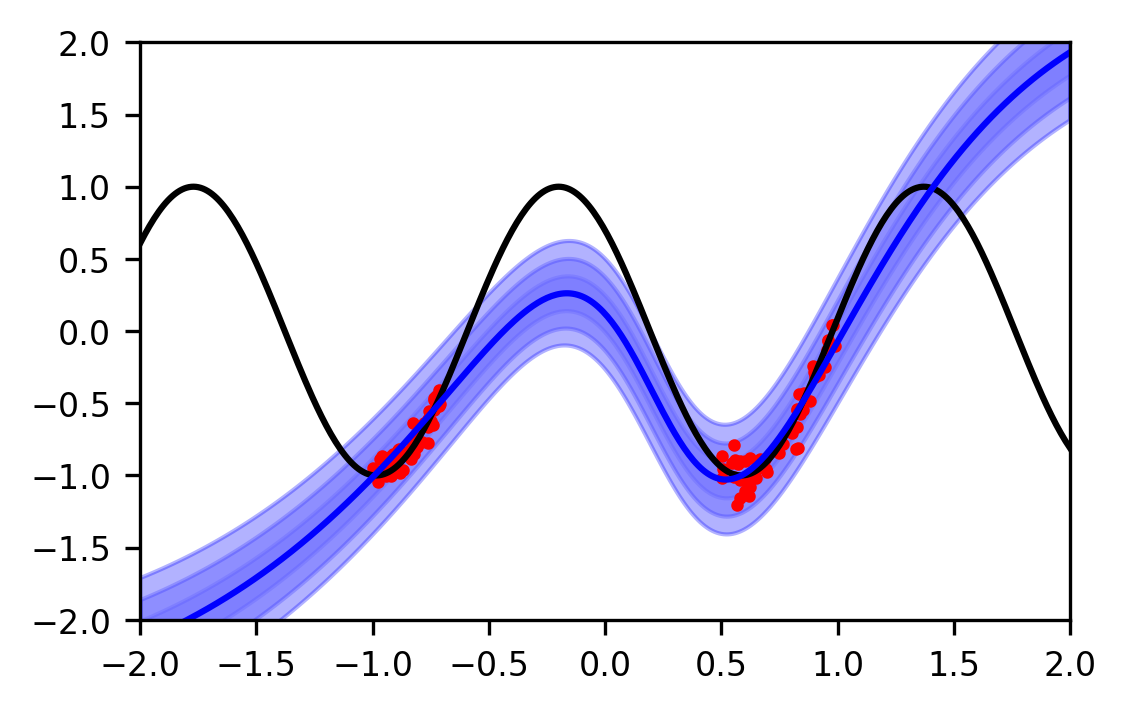}
    }%
    \hfill
    \subfigure[HMC]{\label{subfig:hmc}
        \includegraphics[width=0.3\linewidth]{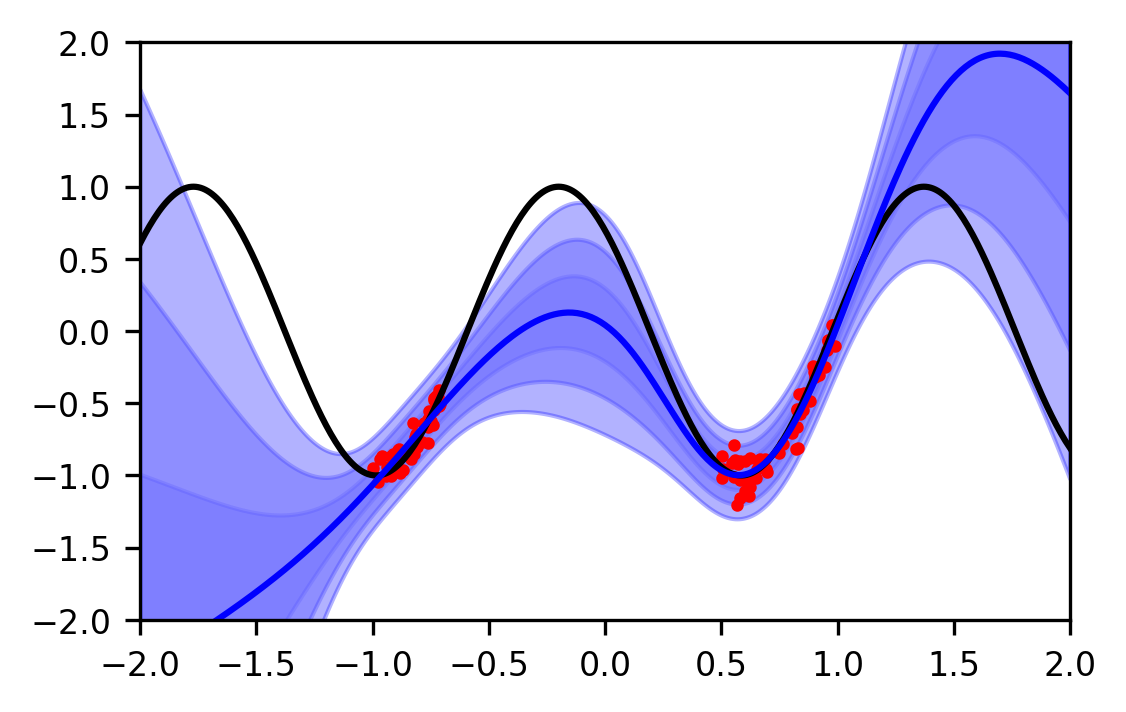}
    }
    \vspace{-0.75em}
    \caption{Bayesian nonlinear regression using the setup from \Cref{lst:bnn} and fit using \Cref{lst:fitpredict}. \Cref{subfig:lr} wraps the call to \inlinepython{bnn.predict} in the local reparameterization context with the call to \inlinepython{fit}, \Cref{subfig:nolr} does not. Switching between the two is as simple as adapting the indentation of the call to \inlinepython{predict} to be in- or outside the \inlinepython{local_reparameterization} context. Both use the same bnn object with the same approximate posterior. \Cref{subfig:hmc} uses \inlinepython{pyro.infer.mcmc.HMC} as guide factory. The shaded area indicates up to three standard deviations from the predictive mean.}
    \label{fig:regression}
    \vspace{-1.25em}
\end{figure*}

A \tyxe BNN has four components: a \pytorch neural network, a data likelihood, a weight prior and a guide\footnote{Following \pyro's terminology we refer to programs drawing approximate posterior samples ``guides''.} factory for the posterior.
We describe their signature and our instantiations below.
As seen in \Cref{lst:bnn}, turning a \pytorch network into a \tyxe BNN requires as little as five lines of code.

\subsubsection{Network architecture}

\pytorch provides a range of classes that facilitate the construction of neural networks, ranging from simple linear or convolutional layers and nonlinearities as building blocks to higher-level classes that compose these, e.g. by chaining them together in the \inlinepython{nn.Sequential} module.
A simple regression network on $1d$ data with one layer of $50$ hidden units and a $\tanh$ nonlinearity, as commonly used for illustration in works on Bayesian neural networks, can be defined in a single line of code (first line of \Cref{lst:bnn}).
More generally, any neural network in \pytorch is described by the \inlinepython{nn.Module} class, which provides functionalities such as easy composition, parameter and gradient handling, and many more conveniences for neural network researchers and practitioners that have contributed to the wide adoption of this framework.
Further, the \inlinepython{torchvision} package implements various modern architectures, such as ResNets \citep{he2016deep}.
TyXe can also work on top of architectures from 3rd party libraries, such as DGL \citep{wang2019dgl}, that derive from \inlinepython{nn.Module}.

\pyro inherits the elegant abstractions for neural networks from \pytorch through its \inlinepython{PyroModule} class, which extends \inlinepython{nn.Module} to allow for instance attributes to be modified by \pyro effect handlers, making it easy to replace \inlinepython{nn.Parameter}s with \pyro sample sites.
We adopt the \inlinepython{PyroModule} class under the hood to provide a seamless interface between \tyxe and \pytorch networks.

\subsubsection{Prior}

At this time, we restrict the probabilistic model definition to weight space priors.
Our classes take care of constructing distribution objects that replace the network parameters as \inlinepython{PyroSample}s.
One such prior class is an \inlinepython{IIDPrior} which takes a \pyro distribution as argument, such as a \inlinepython{pyro.distributions.Normal(0., 1.)}, applying a standard normal prior over all network parameters.
We further implement \inlinepython{LayerwiseNormalPrior}, a per-layer Gaussian prior that sets the variance to the inverse of the number of input units as recommended in \citep{neal1996priors}, or analogous to the variance used for weight initialization in~\citep{glorot2010understanding, he2015delving} when using the flag \inlinepython{method={"radford", "xavier", "kaiming"}}, respectively.
Crucially, we do not require users to set priors for each layer by hand, this is dealt with automatically by our framework.

Our prior classes accept arguments that allow for certain layers or parameters to be excluded from a Bayesian treatment.
The prior in our ResNet example in \Cref{sec:resnet} receives \inlinepython{hide_module_types=[nn.BatchNorm2d]} to hide the parameters of the BatchNorm modules.
Those parameters stay deterministic and are fit to minimize the log likelihood part of the ELBO.

\subsubsection{Guide}

The guide argument is the only place where the initialization of our \inlinepython{VariationalBNN} and \inlinepython{MCMC_BNN} differs.
\inlinepython{tyxe.VariationalBNN} expects a function that automatically constructs a guide for the network weights, e.g. a \inlinepython{pyro.infer.autoguide}, and an optional second such function for variables in the likelihood if present (e.g. an unknown variance in a Gaussian likelihood).

To facilitate local reparameterization and computation of KL-divergences in closed form, we implement an \inlinepython{AutoNormal} guide, which samples all unobserved sites in the model from a diagonal Normal.
This is similar to \pyro's \inlinepython{AutoNormal} autoguide, which constructs an auxiliary joint latent variable with a factorized Gaussian distribution.
Variational parameters can be initialized as for autoguides by sampling from the prior/estimating statistics like the prior median, or through additional convenience functions that we provide, such as sampling the means from distributions with variances depending on the numbers of units in the corresponding layers, akin to how deterministic layers are typically initialized.
This also permits initializing means to the values of pre-trained networks, which is particularly convenient when converting a deep network into a BNN.

The \inlinepython{tyxe.MCMC_BNN} class expects an MCMC kernel as guide, either HMC~\citep{neal2012bayesian} or NUTS~\citep{hoffman2014no}, and runs \pyro's MCMC on the full dataset to obtain samples from the posterior.
For both \inlinepython{BNN} classes, arguments to the guide constructor can be passed via \inlinepython{partial} from Python's built-in \inlinepython{functools} module.
\Cref{lst:resnet} shows an example of this.

\subsubsection{Likelihood}

Our likelihoods are wrappers around \pyro's \inlinepython{distributions}, expecting a \inlinepython{dataset_size} argument to correctly scale the KL term when using mini-batches.
Specifically we provide Bernoulli, Categorical, HomoskedasticGaussian and HeterosketdasticGaussian likelihoods.
Implementing a new likelihood requires a \inlinepython{predictive_distribution(predictions)} method returning a \pyro distribution for sampling.
Further, it should provide a method for calculating an error estimate for evaluation, such as the squared error for Gaussian models or classification error for discrete models.
Hence it is easy to add new likelihoods based on existing distributions, e.g. a Poisson likelihood.

\subsection{Fitting a BNN}

Our BNN class provides a scikit-learn-style \inlinepython{fit} function to run inference for a given numbers of passes over an \inlinepython{Iterable}, e.g. a PyTorch \inlinepython{DataLoader}.
Each element is a length-two tuple, where the first element contains the network inputs (and may be a list) and the second is the likelihood targets, e.g. class labels.
The \inlinepython{VariationalBNN} class further requires a \pyro optimizer.

\inlinepython{tyxe.VariationalBNN} runs stochastic variational inference~\citep{ranganath2014black, wingate2013automated}, a popular training algorithm for Bayesian Neural Networks, e.g.~\citep{blundell2015weight} based on maximizing the evidence lower bound (ELBO).
Our implementation automatically handles correctly scaling the KL-term vs. the log likelihood in the ELBO.
$\inlinepython{tyxe.MCMC_BNN}$ provides a compatible interface to \pyro's \inlinepython{MCMC} class.

\begin{wrapfigure}{l}{.5\textwidth}
\vspace{-1.5em}
\centering
\inputpython{snippets/fit_predict.py}
\vspace{-0.75em}
\captionof{listing}{Regression fit and predict example with local reparameterization enabled for training only.}
\label{lst:fitpredict}
\vspace{-0.75em}
\end{wrapfigure}

\Cref{lst:fitpredict} shows a call to \inlinepython{fit}.
Besides the data loader and number of epochs or samples, it is possible to pass in a \inlinepython{callback} function to the \inlinepython{VariationalBNN}, which is invoked after every epoch with the average value of the ELBO over the epoch and can be used e.g. to check the log likelihood of a validation data set.
By returning \inlinepython{True}, the callback function can stop training.
The \inlinepython{MCMC_BNN} passes any keyword arguments on to \pyro's \inlinepython{MCMC} class.

\subsection{Predicting with a BNN}

The \inlinepython{predict} method returns predictions for a given number of weight samples from the approximate posterior.
\Cref{lst:fitpredict} invokes \inlinepython{predict} at the bottom.
By default it aggregates the sampled predictions, i.e. averages them.
Via \inlinepython{aggregate=False} the sampled predictions can be returned in a stacked tensor.
We further implement an \inlinepython{evaluate} method that expects test labels and returns their log likelihood along with an error measure depending on the model, e.g. squared error for Gaussian likelihoods and classification error for Categorical or Binary ones.

\subsection{Transformations via effect handlers} \label{sec:reparam}

One crucial component missing from \pyro that \tyxe provides is BNN-specific effect handlers~ \citep{plotkin2009handlers,bingham2019pyro}, specifically local reparameterization \citep{kingma2015variational} and flipout \citep{wen2018flipout} for gradient variance reduction.
Local reparameterization samples the pre-activations of each data point rather than a single weight matrix shared across a mini-batch for factorized Gaussian approximate posteriors over the weights and layers performing linear mappings, such as dense or convolutional layers.
Flipout, on the other hand, samples a rank-one matrix of signs per data point, which allows for using distinct weights in a computationally efficient manner in linear operations, if the weights are sampled from a factorized symmetric distribution.

\begin{listing}[t]
\centering
\inputpython{snippets/resnet.py}
\vspace{-0.75em}
\caption{Bayesian ResNet. Line $1$ loads a ResNet with pre-trained parameters from \inlinepython{torchvision}. The prior in lines $2{-}3$ excludes BatchNorm layers, keeping their parameters deterministic. Arguments to the guide are passed with \inlinepython{partial} as in lines $5{-}6$. We show how to set the Gaussian means to the pre-trained weights and only fit the variances, which are initialized to be small. The \inlinepython{BNN} object in line $7$ is constructed exactly the same way as in the regression example. Lines $9{-}11$ show an alternative prior that only applies to the final fully-connected layer alongside a \pyro autoguide.}
\label{lst:resnet}
\vspace{-1.25em}
\end{listing}

Typically, these are implemented as separate layer classes, e.g. \citep{tran2019bayesian}.
This creates an unnecessary redundancy in the code base, since there are now two versions of the same model differing only in sampling approaches for gradient estimation at each linear mapping.
From a probabilistic modeling point of view it is preferable to separate model and inference explicitly to facilitate reuse of models and inference approaches.
Fortunately, \pyro provides an expressive module for effect handling, which we can leverage to modify the computation as required.
Specifically, we implement a \inlinepython{LocalReparameterizationMessenger} which marks linear functions called by \pytorch modules, such as \inlinepython{F.linear}, as effectful in order to modify how linear computations are performed as required.
The Messenger maintains references from samples to their distributions and, when a linear function is called in a \inlinepython{local_reparameterization} context on weights from a factorized Gaussian, samples the output from the Gaussian over the result of the linear mapping.

\Cref{lst:fitpredict} calls \inlinepython{fit} in such a context.
The call to \inlinepython{predict} could be wrapped too, but the purpose of local reparameterization and flipout is to reduce gradient variance.
As they double the computational cost, we omit them for testing.

\section{Large-scale vision classification} \label{sec:resnet}

The biggest advantage resulting from our choice not to implement bespoke layer classes is that implementations of popular architectures can immediately be turned into their Bayesian counterparts.
While implementing the two-layer network from the regression example with Bayesian layers is of course not complicated, writing the code for a modern computer vision architecture, e.g. a ResNet \citep{he2016deep}, is significantly more cumbersome and error-prone.
With \tyxe, users can use the ResNet implementation available through \inlinepython{torchvision} as shown in \Cref{lst:resnet}.
In this example we further highlight the flexibility of \tyxe to only perform inference over some parameters while keeping others deterministic by excluding \inlinepython{nn.BatchNorm2d} layers from a Bayesian treatment.

\begin{figure}[t]
    \centering
    \subfigure{
        \includegraphics[width=\linewidth]{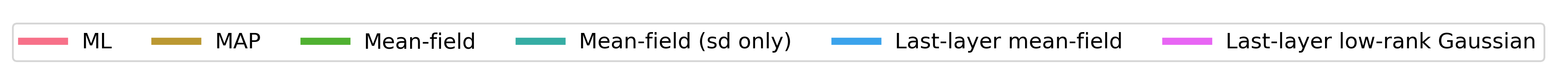}
    }\\[0pt]
    \vspace{-1.25em}
    \setcounter{subfigure}{0}
    \subfigure[Test calibration.]{\label{subfig:calibration}
        \includegraphics[width=0.45\linewidth]{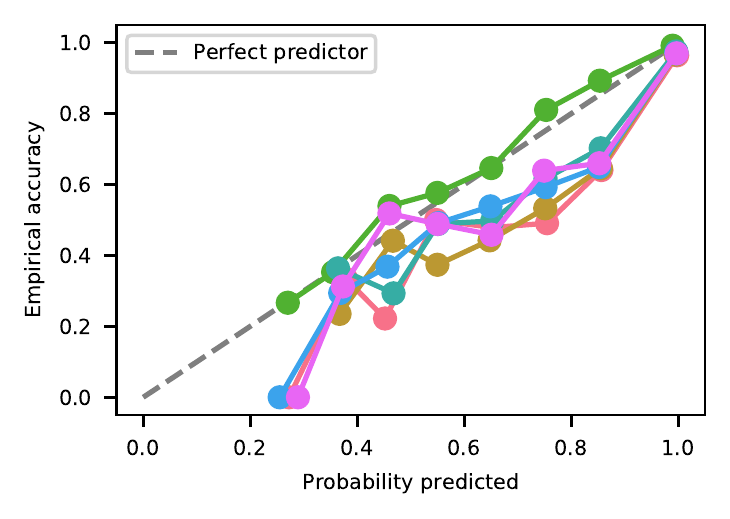}
    }%
    \hfill
    \subfigure[Test vs. OOD uncertainty.]{\label{subfig:ood}
        \includegraphics[width=0.45\linewidth]{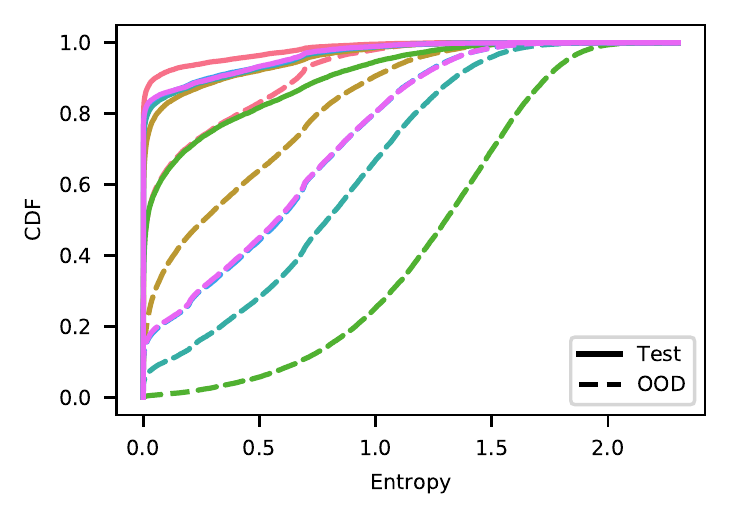}
    }
    \vspace{-0.75em}
    \caption{Calibration curves and empirical cumulative density of the entropy of the predictive distribution on test and OOD data for Bayesian ResNet-18 with different inference approaches on CIFAR10 (OOD: SVHN).}
    \label{fig:bayesian_resnet}
    \vspace{-1.25em}
\end{figure}

To showcase how the clean separation of network architecture, prior, guide and likelihood in \tyxe facilitates an experimental workflow, we investigate the predictive uncertainty of different inference strategies for a Bayesian ResNet.
In \Cref{lst:resnet} we define a fully factorized Gaussian guide that fixes the means to the values of pre-trained weights and only fits the variances as parameters.
While we would usually want the approximate posterior to be as flexible as possible, it has been observed in the literature \citep{louizos2017multiplicative,trippe2018overpruning} that such restrictions can improve the predictive performance of a BNN.
We further investigate a mean-field guide where we similarly initialize the means to pre-trained weight values, but do not fix them for optimization, and restrict the variance of the variational distribution to a maximum of $0.1$ to prevent underfitting.
Finally we test performing inference in only the final classification layer with a Gaussian guide with either a diagonal or low-rank plus diagonal covariance matrix (also shown in the Listing) while using the pre-trained weights for the previous layers.
Switching between these options is easy, with typically only a single or two lines of code differing.
As baselines we compare to maximum likelihood (ML) and maximum a-posteriori (MAP).
For the full code see \inlinepython{examples/resnet.py}.

\begin{wraptable}{r}{0.5\linewidth}
    \centering
    \vspace{-0.75em}
    \caption{Bayesian ResNet-18 predictive perf.}
    \vspace{-0.75em}
    \resizebox{0.9\linewidth}{!}{%
    \begin{tabular}{l|cccc}
        \toprule
        Inference & NLL$\downarrow$ & Acc.$\uparrow$(\%) & ECE$\downarrow$(\%) & OOD$\uparrow$\\
        \midrule
        ML & 0.33 & 94.29 & 4.10 & 0.78 \\
        MAP & 0.29 & 92.14 & 4.44 & 0.82 \\
        MF (sd only) & 0.27 & 93.66 & 3.14 & 0.93 \\
        MF & 0.20 & 93.28 & 0.97 & 0.94 \\
        LL MF & 0.35 & 93.36 & 3.62 & 0.89 \\
        LL low rank & 0.34 & 93.31 & 3.75 & 0.89 \\
        \bottomrule
    \end{tabular}
    }
    \label{tab:resnet}
    \vspace{-0.75em}
\end{wraptable}

\Cref{fig:bayesian_resnet} compares calibration and entropy of the predictive distributions on test and out-of-distribution (OOD) data.
Mean-field (MF) with learned means leads to better calibrated predictions than variants (re-)using point estimates.
It best distinguishes test from OOD data as measured by the area under the ROC curve based on the maximum predicted probability and has the lowest expected calibration error (ECE) and negative log likelihood (NLL), see \Cref{tab:resnet}.

We provide a pure \pyro snippet for a variational ResNet in \Cref{app:comparison} for a direct comparison.
The implementation requires a knowledge of a range of \pyro constructs to avoid pitfalls such as incorrectly scaling prior and likelihood, yet the code ends up being significantly lengthier and somewhat convoluted.
In contrast, \tyxe provides a clean object-oriented interface that will be intuitive for most users with a basic understanding of Bayesian statistics and accessible for pure \pytorch users who do not want to have to learn \pyro.
Crucially, essential features for achieving good discriminative performance with a BNN, such as reparameterization and clipping the variance of the approximate posterior are not available in \pyro.

\section{Compatibility with external libraries}

\tyxe is compatible with libraries outside of the native \pytorch ecosystem and classical settings such as classification of i.i.d. images or regression, as long as the networks build on top of \inlinepython{nn.Module}.
Below, we demonstrate this on a semi-supervised node classification example with a graph neural network from the DGL \citep{wang2019dgl} tutorials, as well as a 3D rendering example in \pytorch{3D}.

\subsection{Bayesian graph neural networks with DGL}
\label{sec:gnn}

\begin{listing}
\centering
\vspace{0.3em}
\begin{minipage}[t]{.49\linewidth}
\inputpython{snippets/gcn_layer.py}
\end{minipage}\hfill%
\begin{minipage}[t]{.49\linewidth}
\inputpython{snippets/gnn_net.py}
\end{minipage}\\[0.1em]
\begin{minipage}[b]{.99\linewidth}
\inputpython{snippets/gnn.py}
\end{minipage}
\vspace{-0.5em}
\captionof{listing}{GNN example. The graph convolutional layer definition (top left) relies on DGL's graph functionality and is used for the GNN (top right). The Bayesian GNN can be constructed in line $1$ with the exact same prior, guide and likelihood options as previously. The \inlinepython{selective_mask} in line $3$ ensures that only predictions on labelled nodes contribute to the log likelihood when calling \inlinepython{fit} in line $4$. The input data now consists of a graph and node features.} 
\label{lst:gnn}
\vspace{-1.25em}
\end{listing}

We extend an example from the DGL tutorials\footnote{\url{https://docs.dgl.ai/en/0.5.x/tutorials/models/1_gnn/1_gcn.html}} to train a Bayesian graph neural network~(GNN) on the Cora dataset.
Graph datasets are often semi-supervised, where an entire graph of nodes is provided, but only some of them are labelled.
Hence we need a mechanism for preventing unlabelled nodes from contributing to the log likelihood.
We combine \pyro's \inlinepython{block} and \inlinepython{mask} poutines to implement the \inlinepython{selective_mask} effect handler, which can wrap the call to \inlinepython{fit} as a context manager as shown in \Cref{lst:gnn} and mask out data in the likelihood.
The network is taken from the DGL tutorial without change.
As it utilizes \inlinepython{nn.Linear}, it is compatible with flipout.
Prior, guide, likelihood and BNN can be constructed exactly as in the previous examples, see \inlinepython{examples/gnn.py} for the code.

\begin{wraptable}{r}{0.5\linewidth}
    \vspace{-1.45em}
    \centering
    \caption{Performance of deterministic and Bayesian GNNs on the Cora dataset. We report the lowest validation NLL along with the test accuracy and ECE at the corresponding epoch (mean and two standard errors over five runs).}
    \vspace{-0.75em}
    \resizebox{0.95\linewidth}{!}{%
    \begin{tabular}{l|ccc}
        \toprule
        Inference & NLL$\downarrow$ & Acc.$\uparrow$ & ECE$\downarrow$ \\
        \midrule
        ML & $1.01 \pm .04$ & $75.64 \pm 1.28$ & $15.38 \pm 0.97$ \\
        MAP & $0.93 \pm .03$ & $75.94 \pm 0.73$ & $12.78 \pm 0.96$ \\
        MF & $0.77 \pm .02$ & $78.02 \pm 1.00$ & $10.22 \pm 1.31$ \\
        \bottomrule
    \end{tabular}
    }
    \label{tab:cora}
\end{wraptable}

In \Cref{tab:cora} we report NLLs, accuracies and ECE for ML, MAP and MF.
ML leads to overfitting and requires the use of early stopping.
Further it suffers from overconfident predictions, which can be mitigated to a degree by the use of variational inference, although not to the same extent as in the image classification example.
Bayesian GNNs have only recently been started to be investigated in a few works \citep{zhang2019bayesian,hasanzadeh2020bayesian,luo2020learning,lamb2020bayesian} and we believe that \tyxe can be a valuable tool for putting Bayesian inference at the disposal of the graph neural network community.

\subsection{Custom losses: Bayesian NeRF with \pytorch{3D}} \label{sec:nerf}

Next, we adapt a more complex example on Neural Radiance Fields (NeRF) \citep{mildenhall2020nerf} from the \pytorch{3D} repository\footnote{\url{https://github.com/facebookresearch/pytorch3d/blob/master/docs/tutorials/fit_simple_neural_radiance_field.ipynb}} to train a Bayesian NeRF.
The loss function does not straight-forwardly correspond to a probabilistic likelihood and is calculated as a custom error function of rendered image and silhouette.
Hence there is no suitable likelihood class to implement for \tyxe and it is not clear how the the prior or KL term should be weighed relative to the error.
Therefore this example is not Bayesian in the proper sense as a `posterior' as a product of likelihood and prior does not exist, but  demonstrates that the uncertainty of a pseudo-Bayesian variational BNN can still improve the robustness on unseen data.

\begin{listing}[t]
\centering
\inputpython{snippets/nerf.py}
\vspace{-0.75em}
\caption{Bayesian NeRF example. Constructing a \inlinepython{PytorchBNN} is similar to a \inlinepython{VariationalBNN} in line $1$ but without the likelihood. No downstream changes except for parameter collection for the \pytorch optimizer in line $2$ -- which requires a batch of data to trace parameters on a call to the net's forward method -- are needed. The \inlinepython{nerf_bnn} can be passed into the \pytorch{3D} renderer in line $4$ as a drop-in replacement for the \inlinepython{nerf_net}. The loss can be calculated as before in lines $5$, with the possible addition of the KL regularizer in line $6$. Automatic differentiation and parameter updates can be performed as in standard \pytorch code in line $7$.}
\label{lst:nerf}
\end{listing}

Specifically, we introduce a more low level \inlinepython{PytorchBNN} class that does not require a likelihood and can be used to directly wrap a \pytorch neural network.
It is constructed similarly to \inlinepython{VariationalBNN} with a variational guide factory, but due to the absence of the likelihood does not provide convenience functions such as \inlinepython{fit} or \inlinepython{predict}.
Instead, it is intended to serve as a drop-in replacement of the deterministic neural network in a \pytorch-based workflow.
The output of the \inlinepython{forward} method corresponds to predictions of the network made with a single Monte Carlo sample from the variational posterior.
The corresponding KL penalty term can be accessed through the \inlinepython{cached_kl_loss} attribute and added to the loss.
It is updated on every forward pass, i.e. when a sample is drawn from the approximate posterior.
The key difference to a regular \pytorch neural network is that since \pyro initializes parameters lazily, we cannot provide a \inlinepython{parameters} method.
Instead, optimizable parameters are collected via \inlinepython{pytorch_parameters}, which takes a batch of data to pass through the network for tracing the parameters.

\begin{wrapfigure}{r}{0.5\textwidth}
    \vspace{-2.75em}
    \centering
    \subfigure{
    \includegraphics[width=0.32\linewidth]{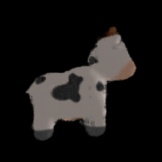}}%
    \hspace{-1em}
    \subfigure{
    \includegraphics[width=0.32\linewidth]{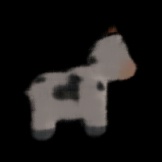}}%
    \hspace{-1em}
    \subfigure{
    \includegraphics[width=0.32\linewidth]{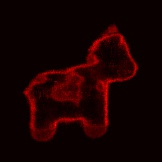}}
    \\[-1.1em]%
    \setcounter{subfigure}{0}
    \subfigure[Det. NeRF]{
    \includegraphics[width=0.32\linewidth]{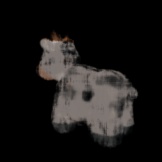}}%
    \hspace{-1em}
    \subfigure[Bay. NeRF]{
    \includegraphics[width=0.32\linewidth]{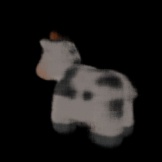}}%
    \hspace{-1em}
    \subfigure[Uncertainty]{
    \includegraphics[width=0.32\linewidth]{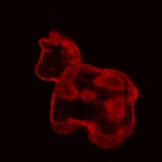}}
    \caption{\pytorch{3D} example. Top row seen during training, bottom row excluded. Bayesian NeRF achieves an error of $8.1{\times}10^{-3}$ on a set of $10$ held-out angles, while the error is $9.4{\times}10^{-3}$ for the deterministic version. Uncertainty visualizes variance across different weight samples.}
    \label{fig:nerf}
    \vspace{-1.25em}
\end{wrapfigure}

We provide a code snippet in \Cref{lst:nerf}.
We emphasize that parameters are trained with the original \pytorch instead of a \pyro optimizer, further reducing the required changes to the original workflow.
The renderer is a \pytorch{3D} object and uses the Bayesian NeRF object instead of the original \pytorch network.
The data-dependent loss is then calculated as before and the KL-divergence of the approximate posterior from the prior on the weights can be added to the objective as a regularizer, possible weighed by some scalar \inlinepython{scale}.
The full code can be found in \inlinepython{examples/nerf.py} and is identical to the original notebook for the most part, with only a few lines needing to be modified to adapt it to \tyxe, as well as some additional plotting code for visualizing the predictive uncertainty.

In the original example, the network is trained to render views of a cow from $360$°.
We hold out $90$° as out-of-distribution data.
As \Cref{fig:nerf} shows, this leads to many artifacts and discontinuities with a deterministic net.
The pseudo-Bayesian NeRF averages many of these out, and provides helpful measures of uncertainty in form of the variances of the predicted images (right column).

\section{Variational continual learning}

\begin{listing}
\centering
\inputpython{snippets/vcl.py}
\vspace{-0.75em}
\caption{Updating the prior of a \inlinepython{BNN} for variational continual learning. Line $1$ collects all weights over which we perform inference, line $2$ extracts the corresponding variational distributions from the guide, and line $3$ uses these to update the BNN's prior.}
\label{lst:vcl}
\vspace{-1.25em}
\end{listing}

Finally, we show how our separation of prior, guide and network architecture enables an elegant implementation of variational continual learning (VCL) \citep{nguyen2017vcl}.
Having set up and trained a BNN on a first task as in the previous examples, we only need construct a new prior from the guide distributions over the weights to update the previous BNN prior.
We show example code for this process in \Cref{lst:vcl} and the full implementation can be found in \inlinepython{examples/vcl.py}.
Training on the following task can then be conducted as usual with the \inlinepython{fit} method on the current dataset.

\begin{wrapfigure}{r}{0.5\textwidth}
    \vspace{-0.75em}
    \centering
    \includegraphics[width=\linewidth]{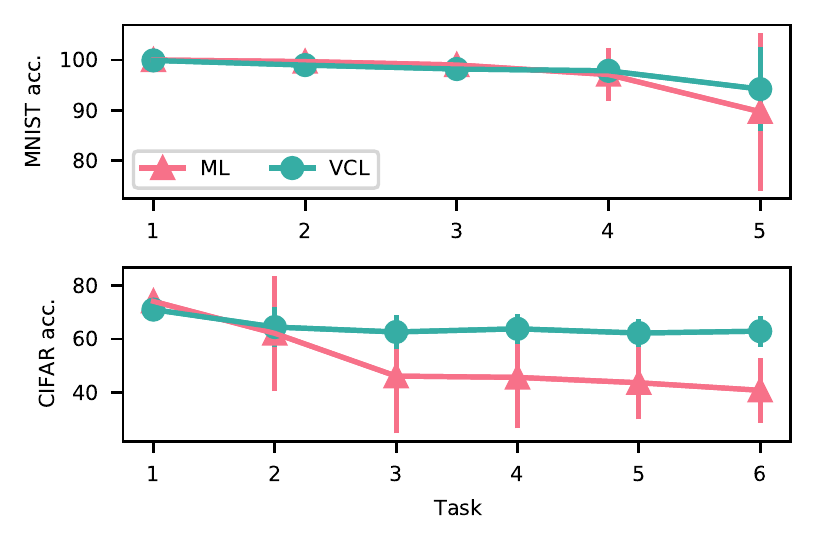}
    \vspace{-2em}
    \caption{Mean accuracy and two standard errors on tasks seen so far for VCL and ML on Split-MNIST and -CIFAR.}
    \label{fig:vcl}
    \vspace{-3.5em}
\end{wrapfigure}

In \Cref{fig:vcl} we show the test accuracy across the observed tasks after training on each one on the classical Split-MNIST and Split-CIFAR benchmarks\citep{zenke2017continual}.
We do not use coresets as \citep{nguyen2017vcl}, but this would only require some boilerplate code for creating the coresets prior to training and then fine-tuning on each coreset prior to testing by calling \inlinepython{fit} and restoring the state of the \pyro parameter store.
As previously reported in the literature, deterministic networks suffer from forgetting on previous tasks, which can be mitigated by using a Bayesian approach such as VCL.

\section{Related work}

The most closely related piece of recent work is Bayesian Layers \citep{tran2019bayesian}, which extends the layer classes of Keras with the aim of them being usable as drop-in replacements for their deterministic counterpart.
This forces the user to modify the code where the network is defined or write their own boilerplate code.
Bayesian Layers are currently more general in scope, providing an abstraction over uncertainty over composable functions including normalizing flows and Gaussian Process mappings per layer, while at this point we have consciously limited ourselves to weight space uncertainty in neural networks and treat networks holistically rather than per layer.

For \pytorch, PyVarInf\footnote{\url{https://github.com/ctallec/pyvarinf}} provides functionality for turning \inlinepython{nn.Module}s into BNNs in a similar spirit to \tyxe.
As it is not backed by a probabilistic programming framework, the choice of prior distributions is limited, inference is restricted to variational factorized Gaussians, sampling tricks such as local reparameterization are not implemented and MCMC-based inference is not available.
More recently, BLiTZ\footnote{\url{https://github.com/piEsposito/blitz-bayesian-deep-learning}} \citep{esposito2020blitzbdl} provides variational counterparts to \pytorch's linear, convolutional and some recurrent layers.
Networks need to be constructed manually based on those, with no support of other layer types.
Priors are limited to mixtures of up to two Gaussians and inference is performed with a factorized Gaussian without support for gradient variance reduction techniques.

Subsequent to release of \tyxe, UQ360 \citep{uq360-june-2021} and BNNPriors \citep{bnnpriors}  were released. BNNPriors provides support for a range of different weight priors, restricting inference to (stochastic) MCMC-based methods; UQ360 provides a general treatment of uncertainty techniques.

\section{Conclusion}

We have presented \tyxe, a \pyro-based library that facilitates a seamless integration of Bayesian neural networks for uncertainty estimation and continual learning into \pytorch-based workflows.
We have demonstrated the flexibility of \tyxe with applications based on 3rd-party libraries, ranging from modern deep image classification architectures over graph neural networks to neural radiance fields.
\tyxe avoids implementing bespoke layer classes and instead leverages and expands on \pyro's powerful effect handler module, resulting in a flexible design that cleanly separates architecture definition, prior, inference, likelihood and sampling logic.

\tyxe's choices of variational distributions are currently pragmatic, focused on serving practitioners and researchers interested in generating uncertainty estimates for downstream tasks that will benefit from the improvements offered by standard variational families or HMC over maximum likelihood.
Recent work has even argued that mean-field may be sufficient for inference in deep networks \citep{farquhar2020liberty}.
However, we are highly interested in further developing \tyxe to support more complex recent approaches and become a tool for Bayesian deep learning research with its backing by Pyro facilitating extensions; see \Cref{app:directions} for an in-depth outlook.
We would expect techniques with structured covariance matrices \citep{louizos2016structured,ritter2018scalable} as well as hierarchical weight models \citep{louizos2017multiplicative,karaletsos2018probabilistic,inducing:weights} to be feasible to express within \tyxe, with the latter possibly requiring additional abstractions.
Nevertheless, we believe that similar to Bayesian Layers \citep{tran2019bayesian} \tyxe already makes a valuable contribution to the ML software ecosystem, filling the gap of easy-to-use uncertainty estimation for \pytorch.


\bibliography{tyxe}
\bibliographystyle{icml2021}

\clearpage
\appendix

\section{Experimental details} \label{app:experiments}

Here we provide hyperparameter details on the experimental results.
Please see the corresponding scripts in the \inlinepython{examples/} directory of the codebase.
In all experiments, we use a single Monte Carlo sample for estimating the expected log likelihood during training for variational inference and standard deviations are initialized to $10^{-4}$ unless stated otherwise.
Weight priors are standard Normals.
Each experiment was run on a single NVIDIA RTX 2080Ti.

\subsection{ResNet}

We use the usual data augmentation techniques for CIFAR10 of randomly flipping and cropping the images after padding them with $4$ pixels on each dimension and we normalize all channels to have zero-mean and unit-standard deviation.
All methods use the Adam optimizer \citep{kingma2014adam}.
We train the deterministic inference methods (ML, MAP) for $200$ epochs with a learning rate of $10^{-3}$ and another $100$ epochs with a learning rate of $10^{-4}$.
All variational methods are trained for $200$ epochs with a learning rate of $10^{-3}$ and we initialize the means to pre-trained ML parameters.
We use a rank of $10$ for the low-rank plus diagonal posterior and average over $32$ samples for predictions on the test and OOD sets.
The factorized Gaussian posteriors all use local reparameterization and we limit the standard deviation of the mean-field posteriors to $0.1$.

\subsection{GNN}

Following the DGL tutorial, we train ML (and MAP) for $200$ iterations with a learning rate of $10^{-2}$ using Adam.
We report the test accuracy at the iteration with lowest validation negative log likelihood.
For mean-field, we train for $400$ iterations with an initial learning rate of $0.1$, which we decay by a factor of $10$ every $100$ iterations and we limit the variational standard deviations to $0.3$.
Means are initialized to the random initialization of the deterministic network and we draw $8$ posterior samples for evaluation.
We use $10$ bins to calculate the expected calibration error.

\subsection{NeRF}

We train the deterministic NeRF with the recommended settings from the tutorial, i.e. $20,000$ iterations with an initial learning rate of $10^{-3}$, which is decayed by a factor of $10$ for the final $5000$ iterations.
The Bayesian NeRF uses the same learning rate schedule.
Means are initialized to the parameters of the deterministic NeRF and standard deviations to $10^{-2}$.
We linearly anneal the weight of the KL term over the first $10,000$ iterations to the inverse of the number of RGB values in the colour images plus the number silhouette pixels.
We use $8$ samples for test predictions and calculate averages and standard deviation over the final image outputs of the renderer.

\subsection{VCL}

Following the recommendations in \citep{swaroop2019improving}, we train on each MNIST task for $600$ epochs and each CIFAR task for $60$.
We use Adam with a learning rate of $10^{-3}$.
The architecture on MNIST is a fully connected network with a hidden layer of $200$ units with ReLU nonlinearities.
The convolutional architecture on CIFAR has two blocks of $Conv-ReLU-Conv-ReLU-Maxpool$ followed by a fully connected layer with $512$ units.
The convolution layers in the first block have $32$, in the second block $64$ channels and all use $3\times3$ kernels with a stride and padding of $1$.
The maxpool operation is $2\times2$ with a stride of $2$.
We normalize all CIFAR tasks to have zero-mean and unit-standard deviation per channel and do not use any form of data augmentation.




\section{Comparison to \pyro}
\label{app:comparison}

\begin{listing}[t]
    \centering
    \inputpython{snippets/pyro_resnet.py}
    \caption{Code snippet for defining and training a pure \pyro variational ResNet.}
    \label{lst:pyro:resnet}
\end{listing}

Here, we show an example of a variational ResNet in pure \pyro and discuss some of the differences to the \tyxe version.
We stress that \Cref{lst:resnet} and \Cref{lst:pyro:resnet} are \textbf{not} equivalent, as \pyro lacks an implementation of local reparemterization and its autoguide classes do not support clipping the variance of a variational guide, so custom implementations of these features would be required in order to achieve competitive discriminative performance.

While \tyxe has an object-oriented interface for constructing prior, likelihood, guide and finally the BNN, with classes for the typical components of a Bayesian modelling workflow, \pyro requires the user to modify their neural network in-place and manually set prior distributions as attributes of the object itself.
The likelihood then requires defining a function that calls the (now Bayesian) network in a probabilistic program to pass into \pyro's SVI class.
Finally, \tyxe removes the need for boilerplate code when training and making predictions.
In particular the latter again requires either familiarity \pyro's poutine library or the Predictive class to perform a standard step of the Bayesian deep learning workflow (which is arguably less common in a smaller-scale Bayesian modelling workflow for which \pyro was primarily designed, where we may be interested in modelling and making inferences on a given dataset rather than predicting on held-out test data).
Overall, we are convinced that \tyxe makes \pyro's functionality significantly more accessible for \pytorch users interested in adding Bayesian methods to their workflow, in particular for those that have been primarily focussed on pure deep learning, as \tyxe has been designed with Bayesian deep learning at its core.
Finally, \tyxe fills critical gaps in the functionality \pyro for successfully training performant BNNs and further provides support for continual learning.

\section{Outline of the current codebase} \label{app:code}

In this section we give an overview of the most relevant modules and classes in our codebase to guide the interested reader through our design and towards aspects of the library that are most relevant to them and discuss some implementation details.

\begin{description}
    \item[\texttt{tyxe/bnn.py}] This module contains our top-level BNN classes. They mostly act as wrappers leveraging the functionality of our \inlinepython{Prior} classes to turn given \pytorch \inlinepython{nn.Module}s into \inlinepython{PyroModule}s and define the probabilistic models to perform inference in. Further, they provide high-level functionality for training, prediction and evaluation, which is delegated to the \inlinepython{Likelihood} classes and \pyro.
    \begin{description}
        \item[\_BNN] Base class for all \tyxe BNNs. Turns a \pytorch neural network into a \inlinepython{PyroModule} given a \inlinepython{Prior} and gives access to a forward pass through the network with samples from the prior.
        \item[GuidedBNN] Base class on top of \inlinepython{_BNN} that gives access to a forward pass with the network given a trace (i.e. sample) from some inference procedure for the network.
        \item[PytorchBNN] Class for constructing objects that can act as variational drop-in replacements of \pytorch network objects in existing codebases. The forward method acts like the forward method of an \inlinepython{nn.Module} object, except that the weights and therefore the function itself are stochastic. Under the hood, as the BNN class has no control over whether the guide function returns the network output (for example, \pyro's AutoGuides do not) it makes the forward pass have the side-effect of caching the output with the object, so that it can be returned. Similarly it stores the KL divergence between approximate posterior and prior (which may be approximated stochastically in each forward pass as the difference of the log densities with the sample from the approximate posterior if the KL is not available in closed form) to use as a regularization term. A further implementation challenge is collecting parameters as in the \inlinepython{.pytorch_parameters} method, which is the equivalent of \inlinepython{.parameters} on an \inlinepython{nn.Module}. As \pyro programs, e.g. the AutoGuides, typically initialize their parameters, a sample batch of data is required to run both prior (which may have parameters to be estimated via maximum likelihood) and posterior and capture all parameters.
        \item[\_SupervisedBNN] Base class inheriting from \inlinepython{GuidedBNN} that now also incorporates a \inlinepython{Likelihood} describing a model for observed data that is conditioned on the output of the neural network. Defines an API for a \inlinepython{predict} method that will run forward passes through the network for multiple posterior samples and either return them as a stacked tensor or aggregate them in a likelihood-dependent way (e.g. average them for predicted class probabilities).
        \item[VariationalBNN] Class for variational Bayesian neural networks in a supervised learning setting. Allows for an additional guide constructor to be passed in for the likelihood if it contains any variables to be inferred (e.g. the unknown variance of a Gaussian observation model). Further provides a \inlinepython{.fit} method that wraps \pyro's \inlinepython{SVI} and \inlinepython{Trace_ELBO} objects to minimize the variational lower bound w.r.t. parameters.
        \item[MCMC\_BNN] Class for MCMC-based BNNs in a supervised learning setting based on \pyro's MCMC kernels. Similarly provides \inlinepython{fit} and \inlinepython{predict} methods that run MCMC on some data and make predictions using those samples respectively.
    \end{description}
    \item[\texttt{tyxe/likelihoods.py}] The likelihood classes are designed as high-level wrapper around \inlinepython{pyro.distributions} that take care of constructing a \pyro function for sampling in a forward, i.e. describing a probabilistic program for the data. Crucially by providing the dataset size, this handles correctly scaling the log likelihood against the log prior (or KL divergence in variational inference) for mini-batches of data. Further they implement logic for evaluating test predictions through the log likelihood and some error measure and aggregating multiple predictions.
    \begin{description}
        \item[Likelihood] Base class for all likelihood classes. Implements all high-level functionality around model construction and evaluation. Expects subclasses to provide functions that construct distribution objects from given network predictions as well handling aggregating multiple predictions and providing an error function.
        \item[\_Discrete] Base class for Bernoulli and Categorical that handles all classification-related logic of error calculation and averaging of predicted probabilities or logits.
        \item[Bernoulli] Likelihood class for binary observations.
        \item[Categorical] Likelihood class for categorical observations.
        \item[Gaussian] Base class for Gaussians. Uses the squared error as the error measures and aggregates predictions to a mean and standard deviation.
        \item[HeteroskedasticGaussian] Gaussian likelihood that assume $2d$ dimensional predictions for $d$ dimensional observation, with the first half of the dimension encoding a mean and the second half the standard deviation. Uses the predicted standard deviations to weigh means according to their precision when aggregating.
        \item[HomoskedasticGaussian]  Gaussian likelihood that assumes a shared variance for all observations. Crucially, this variance may be a probabilistic function that places a prior on an unknown variance in order to support inference over this additional variable.
    \end{description}
    \item[\texttt{tyxe/priors.py}] The prior module provides classes that handle logic around replacing \inlinepython{nn.Parameter} attributes of \inlinepython{nn.Module} objects with \inlinepython{PyroSample} with some prior distribution when turning a network into a \inlinepython{PyroModule}. They further support updating the prior attributes of the \inlinepython{PyroSamples} in order to facilitate continual learning.
    \begin{description}
        \item[Prior] Base class for prior classes that implements apply and update functions for the conversion and updating of \inlinepython{nn.Module}s. Further handles hide/expose functionality (following the logic of \pyro's \inlinepython{block} poutine) that allows excluding/including different network parameters based on their module instance (i.e. a specific layer), their module type (i.e. a specific layer class), their attribute name or their full name. This way entire layers classes can be excluded from a Bayesian treatment, e.g. BatchNorm layers, only specific layers can be considered, e.g. the final layer, or specific parameters can be left to be learned via maximum likelihood estimation, e.g. the bias terms and gives the user high flexibility for their probabilistic model specification through a simple and compact interface.
        \item[IIDPrior] Class for i.i.d. prior across all parameters based on a given distribution object, e.g. \inlinepython{dist.Normal(0, 1)}.
        \item[LayerwiseNormalPrior] Convenience class for per-layer i.i.d. Gaussian priors with variance depending on weight shape, e.g. inversely proportional to the number of input dimensions.
        \item[DictPrior] Convenience wrapper around dictionaries to map parameter names to distributions, e.g. for continual learning.
        \item[LambdaPrior] Convenience wrapper around functions that dynamically generate a distribution for a given parameter object. 
    \end{description}
    \item[\texttt{tyxe/guides.py}] This module builds on top of \pyro's autoguide module. It provides an equivalent to the \textbf{AutoNormal} class that samples from the approximate posterior directly rather than transforming samples and wrapping them in a Delta distribution, order to be compatible with local reparameterization and calculating KL divergences in closed form. Further it provides convenience features such as limiting the variance of the learned posterior or only learning means or variances, which are not generally of interest for Bayesian inference as they reduce rather than increase the flexibility of the variational posterior, however these can improve the discriminative performance of Bayesian neural networks which may outweigh concerns over approximating the true posterior as closely as possible. Finally, the class provides a method for returning a dictionary mapping sampling site names to distributions with detached parameters, which simplifies turning a guide object into a prior. The module further provides some neural-network style initialization functions for variational mean parameters.
    \item[\texttt{tyxe/poutine/reparameterization\_messenger.py}] Effect handlers for reparameterization of certain linear operations. Specifically, these replace samples from weight distributions with samples from the distributions over the outputs, which reduces gradient variance. We implement these operations as `Messenger' classes to be compatible with \pyro's poutine library.
    \begin{description}
        \item[\_ReparameterizationMessenger] Base class that marks \pytorch functions as effectful to register them with \pyro's effect handling stack and handles basic logic around catching reparameterizable sites (currently only fully-factorized Gaussian and Delta distributions). The core idea is to monkey-patch \pytorch functions used by linear layers such as \inlinepython{nn.Linear} and \inlinepython{nn.Conv} with a version of the corresponding \inlinepython{F.linear} and \inlinepython{F.conv} function wrapped in \pyro's \inlinepython{effectful} decorator. This does not modify any behaviour outside of \pyro functions, but allows for subclasses of \inlinepython{Messenger} to modify the behavious of these functions at runtime. We use this to maintain a mapping from samples of tensor to their respective sampling distributions to check if the weights of a reparameterizable function come from a compatible distribution. If that is the case, we delegate sampling the corresponding output to a \inlinepython{_reparameterize} method that is to be implemented in a subclass.
        \item[LocalReparameterizationMessenger] Implements local reparameterization~\citep{kingma2015variational} logic for reparameterization.
        \item[FlipoutMessenger] Implements flipout~\citep{wen2018flipout} logic for reparameterization.
    \end{description}
\end{description}

\section{Future directions} \label{app:directions}

In the long-term we view \tyxe as a high-level BNN interface that complements \pyro with features specific to Bayesian deep learning that are not of interest for more general probabilistic programs.
Below we discuss both some specific components as well as broader directions in which we plan to move \tyxe in the future.

\begin{itemize}
    \item There is a wide range of BNN-specific variational inference approaches that has been developed in the literature over recent years. Some of these, e.g. \citep{swiatkowski2020k} and \citep{tomczak2020efficient} lend themselves particularly well to the abstractions that we have built and we plan to implement these methods. Especially the latter can be conveniently factorized into a general-purpose AutoGuide and logic that extends our local reparameterization effect handlers.
    \item On a related note, we are interested in adding a layer of abstraction to \pyro's AutoGuides that simplifies the construction of AutoGuide classes with some family of distribution that is shared across all variables for infererence. This feature could conveniently provide matrix normal posteriors \citep{louizos2016structured}.
    \item While \pyro provides some full-batch MCMC samplers such as HMC and NUTS, more scalable mini-batch methods are not available, such as SGLD~\citep{sgld}. We intend to add the necessary abstractions and make them available through \tyxe or directly contribute them to \pyro.
    \item Our layer-free implementation of local reparameterization further offers the possibility of formulating and training stochastic binary neural networks \citep{binary:local:reparameterization} in a Bayesian framework, which is an exiting direction of research that we intend to explore.
    \item A pragmatic approach for uncertainty estimation in practice is Monte Carlo Dropout~\citep{dropout:bayesian}. Typical implementations such as in \pytorch implicitly draw a single weight sample per input, however for visualization purposes it can be desirable to fix a single sample across batches of data. Registering Dropout layers as an effect handler could give access to this functionality through \pyro's poutine library. 
    \item To further enhance \tyxe's scope and support research on BNNs in addition to applying them to practical problems, we are highly interested in exploring if moment propagation approaches such as \citep{hernandez2015probabilistic} and \citep{deterministic:vi} can be implemented as effect handlers in a similar spirit to our reparameterization poutines. This may require marking nonlinearity functions as effectful in addition to linear and convolutional layers in order to allow passing distributions rather than tensors through a given network and could greatly facilitate experimenting with such approaches on a broad range architectures with existing \pytorch implementations.
\end{itemize}

\end{document}